\documentclass{article} 
\usepackage{iclr2016_conference,times}
\usepackage{hyperref}
\usepackage{url}

\def \x {\mathbf{x}}

\def \E {\mathbb{E}}

\def \Hk {{\mathcal{H}_{\kappa}}}

\def \H {\mathcal{H}}

\def \R {\mathbb{R}}

\def \bq {\begin{eqnarray}}
\def \eq {\end{eqnarray}}
\def \bqs {\begin{eqnarray*}}
\def \eqs {\end{eqnarray*}}
\def \sign {\mathrm{sign}}
\usepackage{bm}
\newtheorem{lemma}{Lemma}
\newtheorem{theorem}{Theorem}

\usepackage{subfigure}
\usepackage{amsmath, amssymb}
\usepackage{multirow}
\usepackage{paralist,algorithmic,algorithm}   
\usepackage{amsmath,amssymb,amssymb,fancyhdr,graphicx}
\title{Budget Online Multiple Kernel Learning}

\author{Jing Lu, Steven C.H. Hoi\thanks{Corresponding author: http://stevenhoi.org/, SIS, SMU, 80 Stamford Road, Singapore 178902}, Doyen Sahoo\\
School of Information Systems\\
Singapore Management University\\
Singapore\\
\texttt{chhoi@smu.edu.sg}\\
\texttt{jing.lu.2014@phdis.smu.edu.sg}\\
\texttt{doyensahoo.2014@phdis.smu.edu.sg}\\
\And
Peilin Zhao\\
Institute for Infocomm Research (I2R)\\
A*STAR\\
Singapore\\
\texttt{zhaop@i2r.a-star.edu.sg}
}
\begin{document}

\maketitle

\begin{abstract}
Online learning with multiple kernels has gained increasing interests in recent years and found many applications. For classification tasks, Online Multiple Kernel Classification (OMKC), which learns a kernel based classifier by seeking the optimal linear combination of a pool of single kernel classifiers in an online fashion, achieves superior accuracy and enjoys great flexibility compared with traditional single-kernel classifiers. Despite being studied extensively, existing OMKC algorithms suffer from high computational cost due to their unbounded numbers of support vectors. To overcome this drawback, we present a novel framework of Budget Online Multiple Kernel Learning (BOMKL) and propose a new Sparse Passive Aggressive learning to perform effective budget online learning. Specifically, we adopt a simple yet effective Bernoulli sampling to decide if an incoming instance should be added to the current set of support vectors. By limiting the number of support vectors, our method can significantly accelerate OMKC while maintaining satisfactory accuracy that is comparable to that of the existing OMKC algorithms. We theoretically prove that our new method achieves an optimal regret bound in expectation, and empirically found that the proposed algorithm outperforms various OMKC algorithms and can easily scale up to large datasets.
\end{abstract}

\section{Introduction}

Online Multiple Kernel Learning has been successfully used in many real-world applications including classification \citep{hoi2013online}, regression \citep{sahoo2014online}, similarity learning for multimedia search\citep{xia2014online}, and structured prediction \citep{martins2010online}. In contrast to traditional online kernel methods \citep{DBLP:conf/nips/KivinenSW01} where a single kernel function is often chosen either manually or via some intensive cross-validation process, online multiple kernel learning algorithms learn multiple kernel classifiers and their combination simultaneously. This enables multiple kernel learning algorithms to efficiently learn very complicated patterns in an online fashion without the need of choosing the best kernel function prior to the online learning task.

Despite being extensively studied, conventional online multiple kernel learning methods suffer from the curse of kernelization in large-scale applications, that is, the number of support vectors (SV's) often grows linearly with the number of instances received in the online learning process. This not only results in increasing computational cost, but also the growth of memory need which consequently leads to memory overflow in large-scale online applications.


Recent years have witnessed a variety of studies that attempt to make online kernel learning (with a single kernel) scalable by bounding the number of SV's during the online learning process, known as budget online learning. Examples include Randomized Budget Perceptron (RBP) \citep{cavallanti2007tracking}, Forgetron \citep{DBLP:conf/nips/DekelSS05}, Projectron~\citep{orabona2009bounded}, Budget Passive Aggressive (BPA) learning~\citep{wang2010online}, Bounded Online Gradient Descent (BOGD)~\citep{zhao12bogd}, among others. Although budget online learning has been actively studied, very few existing work systematically addresses the similar problem in online multiple kernel learning which is more challenging than online learning with a single kernel.



One straightforward approach to make online multiple kernel learning scalable is to apply the existing budget online learning algorithms to replace the current online learner for each kernel, and allocate a uniform budget for each kernel due to the lack of prior on the quality of individual kernels. Such a naive approach is simple and easy to implement, but falls short in many aspects. First of all, some efficient budget onilne learning algorithms are too simple to achieve satisfactory accuracy (e.g., RBP\citep{cavallanti2007tracking}), while some other algorithms, despite being more effective, are often computationally very intensive for multiple kernel learning (e.g., Projectron~\citep{orabona2009bounded}). Second, because of the varied performances of different kernel functions in practice, using a simple uniform budget allocation is clearly not optimal as one of key successful principles in online the multiple kernel learning process is to focus more in learning with the better quality kernels.



In this paper, we investigate a novel framework of Budget Online Multiple Kernel Classification (BOMKC) and propose a new algorithm termed online Sparse Passive Aggressive learning (SPA) by extending the popular online Passive Aggressive (PA) technique \citep{crammer2006online}.
The key idea of the proposed method is to explore a simple yet effective stochastic sampling strategy for adding SV's, where the probability of an incoming training instance to become a SV is determined by two factors: (i) the loss suffered by the classifier on this instance; and (ii) the historical accuracy of this individual kernel classifier. By limiting the number of SV's, the proposed BOMKC with SPA significantly accelerates the learning process of OMKC while maintaining satisfactory accuracy which is fairly comparable to that of existing non-budget OMKC algorithms. We theoretically prove that the proposed new algorithm not only bounds the number of SV's but also achieves an optimal regret bound in expectation. Finally, we conduct an extensive set of empirical studies which shows that the proposed algorithm outperforms a variety of existing OMKC algorithms and easily scales to large-scale datasets with million instances.

The rest of this paper is organized as follows. Section 2 formulates the problem of Budget OMKC learning and presents the SPA algorithm for BOMKC. Section 3 gives theoretical analysis of the proposed algorithm and achieves the optimal mistake bound in expectation. Section 4 presents the results of our empirical studies, and finally Section 5 concludes this paper.

\section{Budget Online Multiple Kernel Learning}
In this section, we first formulate the problem setting for Online Multiple Kernel Classification (OMKC), and then present the proposed Budget Online Multiple Kernel Classification (BOMKC) with Sparse Passive Aggressive (SPA) learning for classification tasks.

\subsection{Problem Setting and Preliminaries}
 In a typical binary classification task, our goal is to learn a function $f: \R^d \rightarrow \R$ from a sequence of training examples $\{(\x_1,y_1),\ldots,(\x_T,y_T)\}$, where the feature vector $\x_t\in\mathcal X \subset \R^d$ and the class label $y_t\in \mathcal{Y}=\{+1,-1\}$. We use $\hat{y} = \sign(f(\x))$ to predict the class label, and $|f(\x)|$ to measure the classification confidence.

Consider a collection of $m$ kernel functions $\mathcal{K} = \{\kappa_i:\ \R^d \times \R^d \rightarrow \mathbb{R}, i = 1,\dots,m\}$. Each can be a predefined parametric or nonparametric function. MKC aims to learn a kernel-based prediction model by identifying the best linear combination of the $m$ kernels whose weights are denoted by $\bm\theta = (\theta_1,\dots,\theta_m)$. The learning task can be cast into the following optimization:
\bqs\label{batchMKL}
\min_{\bm\theta \in \Delta} \min_{f \in \mathcal{H}_{K (\bm\theta)}}
\frac{1}{2}\|f\|^2_{\mathcal{H}_{K(\bm\theta)}} + C\sum_{t=1}^T \ell(f(\x_t),y_t)
\eqs
where $\Delta = \{ \bm\theta \in \mathbb{R}^{m}_+\ |\bm\theta^T\textbf{1}_{m} = 1\}$, $K(\bm\theta)(\cdot, \cdot) = \sum_{i=1}^m \theta_i\kappa_i(\cdot, \cdot)$, and $\ell(f(\x_t),y_t)$ is a convex loss function that penalizes the deviation of estimating $f(\x_t)$ from observed labels $y_t$. To simplify the discussion, we denote $\ell_t(f) = \ell(f(\x_t),y_t)$ throughout this paper.

The above convex optimization problem of regular batch MKL have been solved by different optimization schemes~\citep{Goenen2011}.
Despite being studied extensively, it still suffers from the common drawbacks of batch learning methods, i.e., not scalable to large-scale applications, expensive in retraining cost for sequential data and not able to adapt to fast-evolving patterns.

To address the challenges faced by batch MKC methods, several algorithms attempt to solve the MKC problem in an online manner \citep{hoi2013online,sahoo2014online} whose updating scheme usually consists of two steps. First learn a set of effective single kernel classifiers $f_t^i\in\H_{\kappa_i}, i=1,...,m$ using existing online kernel learning algorithms (such as kernel Passive Aggressive and kernel Perceptron). Second, learn an effective classifier $f_t(\x)$ by combining these single kernel classifiers:
\vspace{-0.15 in}\bq
f_t(\x)=\sum_{i=1}^m \theta_t^i f_t^i(\x)\eq\vspace{-0.15 in}

where $\theta_t^i\in [0,1]$ is the combination weight of the classifier with respect to $\kappa_i$ at time $t$. The weights can be updated during the learning process by adopting the Hedge algorithm. 

Compared with batch learning methods, these Online Multiple Kernel Classification algorithms enjoy better scalability to large-scale applications and better applicability to sequential data.
Unfortunately, whenever a new instance is misclassified, it will be added to the SV set. The unbounded number of SV's will eventually lead to memory explosion, making it difficult to scale up in practice.

\subsection{Sparse Passive Aggressive Learning for Online Multiple Kernel Learning}

To overcome the critical limitation of existing OMKC algorithms, we aim to explore novel techniques to build scalable and efficient OMKC algorithms. Similar to the existing OMKC algorithms, the proposed algorithm adopts a two-step updating scheme: (i) update the single kernel classifiers individually; (ii) update the weights for combining the classifiers. In the following, we will present the details of our proposed SPA algorithm.

 Our update strategy of each single kernel classifier is generalized from the PA algorithm~\citep{crammer2006online}, 
At the $t$-step, the online hypothesis will be updated:
\vspace{-0.05 in}\bqs f_{t+1}=\min_{f\in\Hk}\frac{1}{2}\|f-f_t\|^2_\Hk+ \eta \ell_t(f)
\eqs
\noindent where $\eta>0$ and 
$\ell_t(f)=[1-y_tf(\x_t)]_+$ is the hinge loss.
This optimization involves two objectives: the first is to keep the new function close to the old one, while the second is to minimize the loss of the new function on the current example.

To limit the number of SV's in each single kernel classifier, we propose a simple yet effective sampling rule which decides if an incoming instance should be a support vector by performing a Bernoulli trial as follows:
\vspace{-0.1 in}\bqs
\Pr(Z_t^i=1)=\rho_t^i,\quad \rho_t^i=\frac{\min(\alpha,\ell_t(f_t^i))}{\beta}
\eqs\vspace{-0.2 in}

\noindent where $Z_t^i\in\{0,1\}$ is a random variable such that $Z_t^i=1$ indicates that a new SV should be added to update the classifier at the $t$-th step, and $\beta \ge \alpha>0$ are parameters to adjust the ratio of support vectors with some given budget. The above sampling rule has two key concerns: (i) The probability of making the $t$-th step update is always less than $\alpha/\beta$, which avoids assigning too high probability on a noisy instance and guarantees that the ratio of support vectors to total received instances is always bounded in expectation. (ii) An example suffering higher loss has a higher probability of being assigned to the support vector set, which maximizes the learning accuracy by adding informative support vectors or equivalently avoid making unnecessary updates.


After obtaining the random variable $Z_t^i$, we will need to develop an effective strategy for updating the classifier. Following the PA
learning principle, we propose the following updating method:
\bq&&\hspace{0.2in}
f_{t+1}^i=\min_{f\in\Hk}P_t^i(f):=\frac{1}{2}\|f-f_t^i\|^2_\Hk+
\frac{Z_t^i}{\rho_t^i}\eta \ell_t(f)\label{eqn:SPA-update1}
\eq
Note that when $\rho_t=0$, then $Z_t=0$ and we set $Z_t/\rho_t=0$. We adopt the above update since it is an unbiased estimation of that of PA update. 
We can derive a closed-form solution as following.
\vspace{-0.1 in}\bqs
f_{t+1}^i(\cdot)=f_t^i(\cdot)+\tau_t y_t\kappa(\x_t,\cdot), ~~\tau_t=\min(\frac{\eta Z_t^i}{\rho_t^i}, \frac{\ell_t(f_t^i)}{\kappa(\x_t,\x_t)})
\vspace{-0.2 in}\eqs
The remaining problem is to learn the appropriate combination weights for the set of single kernel classifiers. The simplest way is a uniform combination, i.e. $\theta^i=\frac{1}{m}$, which will be used in our initialization step. Hedge algorithm is used to update the weights of each classifier.

In addition, to focus on the best kernel, a larger update probability is assigned to a component classifier with higher historical accuracy, which is reflected by the current combination weight. Finally, we summarize the proposed Sparse Passive Aggressive algorithm in Algorithm~\ref{alg:SPA}


\begin{algorithm}[hpt]
\caption{Sparse Passive Aggressive learning for Budget OMKC ``{\bf BOMKC(SPA)}"}\label{alg:SPA}
\begin{algorithmic}
\STATE {\textbf{INPUT}:}
 Kernels: $k_i(\cdot, \cdot): \mathcal{X}\times\mathcal{X} \rightarrow \mathbb{R}, i=1, \ldots, m$;
 Aggressiveness parameter $\eta>0$, and parameters $\beta\ge \alpha>0$;
 Discount parameter $\gamma\in(0,1)$ and smoothing parameters $\delta_t\in (0,1)$.
\STATE \textbf{Initialization}: $f_1^i =0,  w_1^i=\theta_1^i=\frac{1}{m},i=1,...m$
\STATE \textbf{for} $t = 1,2,\ldots,T$ \textbf{do}
\STATE  \qquad Receive an instance: $\x_t$ and predict $\hat{y}_t = \sign\Big(\sum_{i=1}^{m}\theta_t^i\cdot f_t^i(\x_t)\Big)$
\STATE  \qquad Receive the class label: $y_t$
                       \STATE  \qquad \textbf{for} $i=1, 2, \ldots, m$ \textbf{do}
                       \STATE \qquad \qquad  Compute $p_t^i=(1-\delta_t)\frac{w_t^i}{\max_j w_t^j}+\delta_t$; Bernoulli Sampling $c_t^i\in\{0,1\}$ by $\Pr(c_t^i=1)=p_t^i$\vspace{-0.1 in}
                       \STATE \qquad \qquad  \textbf{if} $c_t^i=1$
                       \STATE \qquad \qquad  \qquad Compute $\rho_t^i = \frac{\min(\alpha, \ell_t(f_t^i))}{\beta}$; Sample $Z_t^i\in\{0,1\}$ by:  $\Pr(Z_t^i=1)=\rho_t^i$
                       \STATE \qquad \qquad  \qquad Update the classifier as Proposition 1
                       \STATE \qquad \qquad Update weight $w_{t}^i=w_t^i\gamma^{\ell_t(f_t^i)}$
\STATE \qquad Scale the weights $\theta_{t+1}^i=\frac{w_{t}^i}{\sum_{j=1}^m w_{t}^j}, i=1,...,m$
\end{algorithmic}
\end{algorithm}\vspace{-0.2 in}
\section{Theoretical Analysis}
In this section, we will provide detailed theoretical analysis to the loss bound of our proposed SPA algorithm. We first propose a lemma that bounds the regret of a single kernel classifier learnt by the SPA update strategy and then utilize this lemma for the final analysis of our proposed SPA algorithm for OMKC problem. We denote
$f_*=\arg\min_{f\in\kappa}\sum^T_{t=1}\ell_t(f)$
as the optimal classifier in $\H_{\kappa}$ space with the assumption of the foresight to all the instances. 

\begin{lemma}\label{thm:expected-regret}
Let $(\x_1,y_1),\ldots,(\x_T,y_T)$ be a sequence of examples where $\x_t\in\R^d$, $y_t\in\mathcal{Y}=\{-1,+1\}$ for all $t$. If we assume $\kappa(\x,\x)=1$ and the hinge loss function $\ell_t(\cdot)$ is $1$-Lipschitz, then for any $\beta\ge\alpha>0$, and $\eta>0$, when using the proposed SPA update strategy to generate a sequence of single kernel classifiers in space $\kappa$, we have
\begin{small}
\vspace{-0.1 in}\bqs
\E[\sum^T_{t=1}(\ell_t(f_t)-\ell_t(f_*))]<\frac{1}{2\eta}\|f_*\|^2_\Hk+ \frac{\eta \beta}{\min(\alpha,\sqrt{\beta\eta})}T
\eqs where $\eta$ is the aggressiveness parameter. Setting $\eta=\|f_*\|_\Hk\sqrt{\frac{\alpha}{2\beta T}}$ and $\alpha^3\le\frac{\beta}{2T}\|f_*\|^2_\Hk$, we have
\vspace{-0.15 in}\bqs
\E[\sum^T_{t=1}(\ell_t(f_t)-\ell_t(f_*))]< \|f_*\|_\Hk \sqrt{2\beta T/\alpha}\vspace{-0.1 in}
\eqs
\end{small}
\end{lemma}
\vspace{-0.1 in}

\noindent{\bf Remark 1.} The theorem indicates that the expected regret of any single classifier in $\H_\kappa$ by the SPA algorithm can be bounded by $\|f_*\|_{\H_{\kappa}} \sqrt{2\beta T/\alpha}$ in expectation. In practice, $\beta/\alpha$ is usually a small constant. Thus the proposed algorithm achieves a strong regret bound in expectation.

\noindent{\bf Remark 2.} The expected regret has a close relationship with the two parameters $\alpha$ and $\beta$. As indicated above, to avoid assigning too high probability on a noisy instance, the parameter $\alpha$ should not be too large. Assuming $\alpha\le \sqrt{\beta\eta}$ (which is accessible in the practical parameter setting), the expected regret bound is proportional to the ratio $\beta/\alpha$. This is consistent with the intuition that larger chances of adding SV's leads to smaller loss. Further more, for $\alpha>\sqrt{\beta\eta}$, the expected regret bound is less tight than the above case, which is consistent with the analysis before that too large $\alpha$ involves large number of noisy instances and might be harmful.

We have demonstrated that the loss of each single kernel online classifier is bounded by the loss of its batch counterpart and an online regret term ($\sqrt{T}$). While in reality, the performance of single kernel classifiers varies significantly and we have no foresight to the winner.
In the following, we will show that the final multiple kernel classifier in our proposed algorithm based on single kernel SPA classifiers achieves almost the same performance as that of the best online single kernel classifier.

\begin{theorem}\label{thm:expected-regret-multi}
Let $(\x_1,y_1),\ldots,(\x_T,y_T)$ be a sequence of examples where $\x_t\in\R^d$, $y_t\in\mathcal{Y}=\{-1,+1\}$ for all $t$. If we assume $\kappa_i(\x,\x)=1$ and the hinge loss function $\ell_t(\cdot)$ is $1$-Lipschitz, then for any $\beta\ge\alpha>0$, and $\eta>0$, $\gamma\in(0,1)$, the multiple kernel classifier generated by our proposed SPA algorithm satisfies the following inequality
\vspace{-0.05 in}\bqs
\E[\sum_{t=1}^T\ell_t(f_t)]\leq \frac{\min_i(\sum^T_{t=1}\ell_t(f_*^i)+\frac{1}{2\eta\delta}\|f_*^i\|^2_\Hk)\mu\ln(1/\gamma)}{1-\gamma}
+ \frac{\mu\eta \beta\ln(1/\gamma)}{\min(\alpha,\sqrt{\beta\eta})(1-\gamma)\delta}T+\frac{\mu\ln m}{1-\gamma}
\eqs
where $\eta$ is the aggressiveness parameter, $\mu$ is a constant that only depends on $\gamma$ and the upper bound of $\ell_t$. Setting $\eta=\|f_*^i\|_\Hk\sqrt{\frac{\alpha}{2\beta T}}$ and $\alpha^3\le\frac{\beta}{2T}\|f_*^i\|^2_\Hk$, we have
\bqs\begin{aligned}
\E[\sum_{t=1}^T\ell_t(f_t)]\leq \frac{\mu\ln m}{1-\gamma}+\frac{\mu\ln(1/\gamma)}{1-\gamma}\min_i\big\{\sum^T_{t=1}\ell_t(f_*^i)+\frac{1}{\delta}\|f_*^i\|_\Hk \sqrt{2\beta T/\alpha}\big\}
\end{aligned}\eqs
\end{theorem}
\vspace{-0.1 in}
\noindent{\bf Remark.} This theorem suggests that even without any foresight of which kernel would achieve the highest accuracy, the performance of our proposed multiple kernel classifier is still comparable to the best single kernel classifier. Compared with the optimal batch classifier with foresight to all the training instances, the regret of our proposed algorithm is $O(\sqrt{T})$.

Next, we will bound the number of SV's of each single kernel classifier $f_T^i$ in expectation.

\begin{theorem}
Let $(\x_1,y_1),\ldots,(\x_T,y_T)$ be a sequence of examples where $\x_t\in\R^d$, $y_t\in\mathcal{Y}=\{-1,+1\}$ for all $t$. If we assume $\kappa(\x,\x)=1$ and the hinge loss function $\ell_t(\cdot)$ is $1$-Lipschitz, $\beta\ge\alpha>0$, and $\eta>0$, for any $i=1,...m$ the proposed SPA algorithm satisfies the following inequality
\begin{small}
\vspace{-0.1 in}\bqs
\begin{aligned}
\E[\sum^T_{t=1}Z_t^i]\hspace{-0.05in}\le\min\left\{\frac{\alpha}{\beta}T, \frac{1}{\beta}[\sum^T_{t=1}\ell_t(f_*^i)+ \frac{1}{2\eta}\|f_*^i\|^2_{\H_{\kappa_i}}+ \frac{\eta \beta}{\min(\alpha,\sqrt{\beta\eta})}T ]\right\}
\end{aligned}
\eqs\end{small}
\noindent Especially, when $\eta=\|f_*^i\|_{\H_{\kappa_i}}\sqrt{\frac{\alpha}{2\beta T}}$ and $\alpha^3\le\frac{\beta}{2T}\|f_*^i\|^2_\Hk$, we have\begin{small}
\vspace{-0.1 in}\bqs
\E[\sum^T_{t=1}Z_t^i]\le \min\left\{\frac{\alpha}{\beta}T, \frac{1}{\beta}\large[\sum^T_{t=1}\ell_t(f_*^i)+ \|f_*^i\|_{\H_{\kappa_i}} \sqrt{2\beta T/\alpha}\large]\right\}
\eqs\end{small}
\end{theorem}\vspace{-0.1 in}
\noindent{\bf Remark.} First, this theorem indicates the expected number of support vectors is less than $\alpha T/\beta$. Thus, by setting $\beta\ge\alpha T/B$ ($1<B\le T$), we guarantee the expected number of support vectors of the final classifier is bounded by a budget $B$. Second, this theorem also implies that, by setting $\beta\ge [\sum^T_{t=1}\ell_t(f_*^i)+ \|f_*^i\|_{\H_{\kappa_i}} \sqrt{2\beta T/\alpha}]/B$ ($1<B\le T$), the expected number of support vectors is always less than $B$, no matter what is the value of $\alpha$. In practice, as the Hedge algorithm trends to converge to a single best kernel, the total number of support vectors of all classifiers is only slightly larger than that used by the best classifier.

\section{Experiments}
In this section, we conduct extensive experiments to evaluate the empirical performance of the proposed SPA multiple kernel algorithm for online binary classification tasks.

\subsection{Experimental Testbed}
All datasets used in our experiments are commonly used benchmark datasets for binary classification. These datasets are chosen
fairly randomly to cover a variety of different sizes with the instances number varies from 1000 to 1,000,000. There details are shown in the Appendix.
\subsection{Kernels}
In our experiments, we examine BOMKC by exploring a set of 16 predefined kernels, including 3 polynomial kernels $\kappa(\x_i,\x_j)=(\x_i^\top\x_j)^p$ with the degree parameter $p=1,2,3$; 13 Gaussian kernels $\kappa(\x_i,\x_j)=\exp(-\frac{||\x_i-\x_j||^2_2}{2\sigma^2})$ with the kernel width parameter $\sigma=[2^{-6},2^{-5},...,2^6]$.
\subsection{Compared Algorithms}
First, we include an important baseline algorithm showing the best performance that could be achieved by a single kernel classifier assuming the foresight of optimal choice of the kernel function before the training instances arrives. We search for the best single kernel classifier from the set of our predefined 16 kernels using one random permutation of all the training examples and then apply the ``Perceptron" algorithm \citep{Rosenblatt58} with the best kernel function.

Our second group of compared algorithms are the Online Multiple Kernel Classification algorithms \citep{hoi2013online} which achieved state-of-the-art performance on many benchmark datasets. Three variants of this algorithm are included:
\begin{itemize}\vspace{-0.1 in}
\item ``OMKC(U)'':  the OMKC algorithm with a naive uniform combination;\vspace{-0.05 in}
\item``OMKC(DD)": the OMKC algorithm with deterministic combination and update;\vspace{-0.05 in}
\item``OMKC(SD)": OMKC with  stochastic update and deterministic combination;\vspace{-0.1 in}
\end{itemize}
Finally, to test the efficiency and effectiveness of our budget strategy, we also compare with multiple kernel classification algorithms whose component classifiers are updated by budget kernel learning algorithms including:
\begin{itemize}\vspace{-0.1 in}
\item ``RBP": the Random Budget Perceptron algorithm \citep{cavallanti2007tracking};\vspace{-0.05 in}
\item ``Forgetron": the Forgetron algorithm that discards the oldest SV \citep{DBLP:conf/nips/DekelSS05};\vspace{-0.05 in}
\item ``BOGD": the Budget Online Gradient Descent algorithm \citep{zhao12bogd};\vspace{-0.05 in}
\item ``BPAS": the Budget Passive-aggressive algorithm (simple)\citep{wang2010online}.\vspace{-0.1 in}
\end{itemize}
Multiple Kernel classification is a more challenging problem compared to single kernel classification, which requires highly efficient component classifiers. Consequently, we did not compare with slow budget algorithm such as Projectron.

\subsection{Parameter Settings}
To make a fair comparison, we adopt the same experimental setup for all the algorithms. The weight discount parameter $\gamma$ is fixed to 0.99 for all multiple kernel algorithms on all datasets. The smoothing parameter $\delta$ for all stochastic update algorithms is fixed to $0.001$. The learning rate parameters in all algorithms (SPA, the BOGD and BPAS) are all fixed to 0.1. For the proposed SPA algorithm, we simply fix the random sampling parameter $\alpha=1$ and $\beta=3$ for the first 8 datasets and will discuss the parameter sensitivity later. For a fair comparison between the budget algorithms, we set
a uniform limit $B$ to the SV size for in all the component classifiers so that the total number of SV's used by all component classifiers $16B$ is nearly equal to the SV size of the SPA algorithm. To test the scalability of our proposed algorithm, we also include the experiment on million scale dateset. We set the $\beta=300$ for efficiency.

All experiments were repeated 10 times on different random permutations of instances and all the results were obtained by averaging over 10 runs. All the algorithms were implemented in C++ on a Windows machine with 3.2 GHz CPU. We report the online mistake rates along the training process, the total number of SV's used by all component single kernel classifiers and the running time.

\subsection{Evaluation of Online Learning Performance with Comparison to Non-Budget OMKL Algorithms}

The first experiment is to evaluate the performance of SPA for binary classification tasks with comparison to non-budget OMKL algorithms.  We did not report the results on the 3 largest datasets since it is nearly impossible for non-budget algorithms to process large scale data in limited time and memory. Table~\ref{tab:1} summarizes the experimental results. We can draw the following observations.

\begin{table*}[ht]\vspace{-0.1in}
\begin{center}
\begin{small}
\caption{\small {Evaluation of Online Multiple Kernel Classification on small-scale and medium-scale datasets.}}\label{tab:1}
\begin{tabular}{l|crr|crr}
\hline
& \multicolumn{3}{|c|}{ german}&
\multicolumn{3}{|c}{ madelon}\\
{Algorithm} &{\scriptsize{Mistake} (\%)} &\scriptsize{\#SV's}&\scriptsize{Time (s)}&\scriptsize{Mistake (\%)}&\scriptsize{\#SV's}&\scriptsize{Time (s)
}\\ \hline
\scriptsize{Perceptron(*)}&32.07 $\pm$	0.90&	320.7 $\pm$	9.0	&0.031     &48.65	$\pm$ 1.57&	973.0	$\pm$31.45&	0.81\\\hline
{OMKC(U)}&35.37 $\pm$	1.17&	6912.4 $\pm$	87.7&	0.388  &50.00	$\pm$ 0.00	&26498.2	$\pm$ 92.62&	35.27\\
{OMKC(DD)}&31.05 $\pm$	1.13&	6912.4 $\pm$	87.7&	0.383  &41.15	$\pm$ 0.50&	26498.2	$\pm$ 92.62&	35.46\\
{OMKC(SD)}&34.80 $\pm$	1.25&	5724.5 $\pm$	128.6&	0.341  &50.00	$\pm$ 0.00	&13872.3	$\pm$ 150.61&	20.67\\

{BOMKC(SPA)}&\textbf{30.19 $\pm$	0.29}&	1688.1 $\pm$	90.7&	\textbf{0.120}&\textbf{36.62	$\pm$ 0.61}	& 8685.2	$\pm$ 93.5	&\textbf{11.7}\\
\hline
\hline
& \multicolumn{3}{|c|}{ a9a}&
\multicolumn{3}{|c}{ magic04}\\
{Algorithm} &{\scriptsize{Mistake} (\%)} &\scriptsize{\#SV's}&\scriptsize{Time (s)}&\scriptsize{Mistake (\%)}&\scriptsize{\#SV's}&\scriptsize{Time (s)
}\\ \hline
\scriptsize{Perceptron(*)}&20.43	$\pm$ 0.13&	\scriptsize{9980.1	$\pm$ 67.4}	&22.5    &24.29	$\pm$ 0.13	& \scriptsize{4620.9	$\pm$ 25.9}	&1.94\\\hline
{OMKC(U)}&20.60	$\pm$ 0.10&	\scriptsize{234008.7	$\pm$ 373.8}	&1178.5 &28.70	$\pm$ 0.12	& \scriptsize{157922.7	$\pm$ 164.4}	&199.5\\
{OMKC(DD)}& 19.20	$\pm$ 0.12&	\scriptsize{234008.7	$\pm$ 373.8} &	1171.6&22.58	$\pm$ 0.46	&\scriptsize{ 157922.7	$\pm$ 164.4}	&199.0\\
{OMKC(SD)}&18.93	$\pm$ 0.10&	\scriptsize{ 155393.9	$\pm$ 560.9}	&778.9&22.39	$\pm$ 0.11	&\scriptsize{ 73433.4	$\pm$ 320.1}	&69.4\\

{BOMKC(SPA)}      &\textbf{16.10	$\pm$ 0.12}&	\scriptsize{7092.3	$\pm$ 328.0} &	\textbf{22.1}&\textbf{19.81	$\pm$ 0.19}	& \scriptsize{4062.9	$\pm$ 235.5}	&\textbf{2.58}\\
\hline
\hline
&
\multicolumn{3}{|c|}{ KDD08}&
\multicolumn{3}{|c}{ svmguide3}
\\
{Algorithm} &\scriptsize{Mistake (\%)} &\scriptsize{\#SV's}&\scriptsize{Time (s)}&\scriptsize{Mistake (\%)}&\scriptsize{\#SV's}&\scriptsize{Time (s)}\\ \hline
\scriptsize{Perceptron(*)}&0.94	$\pm$ 0.01	& 963.5	$\pm$ 11.0&	14.7&25.98	$\pm$ 0.51	&323	$\pm$ 6.3	&0.03\\\hline
{OMKC(U)}&0.66	$\pm$ 0.01	&32665.1	$\pm$ 199.7&	829.7&28.25	$\pm$ 0.80	& 6166.9	$\pm$ 100.1	&0.36\\
{OMKC(DD)} &0.72	$\pm$ 0.01&	32665.1	$\pm$ 199.7&	819.0& 25.41	$\pm$ 0.95	&6166.9	$\pm$ 100.1	&0.37\\
{OMKC(SD)}&0.63	$\pm$ 0.01&	17218.5	$\pm$ 95.3&	450.8&24.65	$\pm$ 0.46	&5550.8	$\pm$ 95.8	& 0.36\\

{BOMKC(SPA)}      &\textbf{0.61	$\pm$ 0.00}	&1687.7	$\pm$ 86.9&	\textbf{34.8}&\textbf{23.53	$\pm$ 0.24}	&1663	$\pm$ 109.63	& \textbf{0.13}\\
\hline
\end{tabular}
\end{small}
\end{center}
\vspace{-0.3in}
\end{table*}

First of all, we compare the accuracy of the three algorithms with deterministic update and full SV's (Perceptron, OMKC(U), OMKC(DD)). Obviously, the mistake rate of OMKC(DD) is usually much lower than that of OMKC(U), which indicates that the OMKC(DD) algorithm is able to learn the best combination to build an effective multiple kernel classifier. Further, to our surprise, even with the unrealistic assumption that we have foresight to all the instances and can choose the best kernel before learning, the Perceptron algorithm still can not achieve the lowest error rate. We conjecture that there might be two explanations to this observation. First, our optimal kernel is searched in one random permutation, which might not be the optimal kernel for other permutations of instances, while OMKC(DD) always learns the best combination of the kernel functions. Second, in some datasets, no single kernel function has significant advantage over others, while an optimal weighted combination might perform better than any single kernel classifier. This further validates the significance of studying multiple kernel learning algorithms.

Second, we find that although using fewer SV's, the stochastic update algorithms OMKC(SD) can even achieve lower mistake rate compared with deterministic update OMKC(DD), which is consistent with the previous observations \citep{hoi2013online}. This indicates that not all SV's are essential for an accurate classifier. This supports our main claim that when the added SV's are wisely selected, the accuracy may not decrease much while the efficiency can be significantly upgraded.

Third, when comparing the time cost, we find that the non-budget OMKC algorithms are significantly slower than our proposed SPA algorithm, which is more serious in larger datasets. Actually, it is even impossible for non-budget algorithms to complete the learning process in realistic time and space in the three largest datasets. This is due to two reasons: first, similar to the problem faced by single kernel methods, the prediction cost grows rapidly with unlimited number of SV's added; second, there is a big pool of kernel classifiers need to be updated and combined. This observation further validates the importance of studying efficient and scalable online multiple kernel methods.

Finally, we found that apart from the obvious advantage in time cost, our proposed SPA algorithm achieves the highest accuracy in all datasets even when adopting only a small portion of SV's. This validates the effectiveness of our proposed algorithm in wisely selecting the most informative SV's.
\subsection{Evaluation of Online Learning Performance with Comparison to Different Budget OMKL Algorithms}
The next experiment is to test the accuracy and efficiency of our proposed SPA algorithm with comparison to other budget maintenance strategies. Table~\ref{tab:2} summarizes the experimental results.

\begin{table*}[ht]
\begin{center}
\begin{small}
\caption{\small {Evaluation of BOMKC with different budget learning algorithms on large-scale datasets.}}\label{tab:2}
\begin{tabular}{l|ccr|ccr}
\hline
BOMKC& \multicolumn{3}{|c|}{\footnotesize a9a}&
\multicolumn{3}{c}{\footnotesize magic04}\\
{Algorithms} &{\scriptsize{Mistake} (\%)} &\scriptsize{\#SV's}&\scriptsize{Time (s)}&\scriptsize{Mistake (\%)}&\scriptsize{\#SV's}&\scriptsize{Time (s)
}\\ \hline
{RBP}      &20.46 $\pm$	0.24&	7088	$\pm$ 0	&38.2&27.90	$\pm$ 0.40	&4064	$\pm$ 0	&4.56\\
{Forgetron}&20.78 $\pm$	0.35&	7088	$\pm$ 0	&54.9&28.20	$\pm$ 0.48	&4064	$\pm$ 0	&7.04\\
{BOGD}     &19.27 $\pm$	0.08&	7088	$\pm$ 0	&42.1&25.68	$\pm$ 0.28	&4064	$\pm$ 0	&5.25\\
{BPAS}     &17.09 $\pm$	0.10&	7088	$\pm$ 0	&42.4&22.73	$\pm$ 0.29	&4064	$\pm$ 0	&5.11\\

{SPA}      &\textbf{16.10	$\pm$ 0.12}&	7092.3	$\pm$ 328.0 &	\textbf{22.1}&\textbf{19.81	$\pm$ 0.19}	&4062.9	$\pm$ 235.5	&\textbf{2.58}\\
\hline
\hline
BOMKC&
\multicolumn{3}{|c|}{\footnotesize KDD08}&
\multicolumn{3}{c}{\footnotesize ijcnn1}
\\
{Algorithms} &\scriptsize{Mistake (\%)} &\scriptsize{\#SV's}&\scriptsize{Time (s)}&\scriptsize{Mistake (\%)}&\scriptsize{\#SV's}&\scriptsize{Time (s)}\\ \hline
{RBP}      &0.838 $\pm$	0.035&	1680	$\pm$ 0	&49.5                        &  8.75	$\pm$ 0.76	&5648	$\pm$ 0	&18.4   \\
{Forgetron}&0.943	$\pm$ 0.036&	1680	$\pm$ 0	&52.2       &  8.74	$\pm$ 0.59	&5648	$\pm$ 0	&21.6   \\
{BOGD}     &\textbf{0.610	$\pm$ 0.000}	&1680	$\pm$ 0	&52.8          &  9.71	$\pm$ 0.01	&5648	$\pm$ 0	&21.6  \\
{BPAS}     &\textbf{0.610	$\pm$ 0.000}	&1680	$\pm$ 0	&52.6        &  9.68	$\pm$ 0.04	&5648	$\pm$ 0	&21.3\\

{SPA}      &\textbf{0.610	$\pm$ 0.000}	&1687.7	$\pm$ 86.9&	\textbf{34.8}&\textbf{7.75	$\pm$ 0.61}&	5653	$\pm$ 369.2&	\textbf{12.9}\\
\hline
\hline
BOMKC&
\multicolumn{3}{|c|}{\footnotesize codrna}&
\multicolumn{3}{c}{\footnotesize SUSY}
\\
{Algorithms} &\scriptsize{Mistake (\%)} &\scriptsize{\#SV's}&\scriptsize{Time (s)}&\scriptsize{Mistake (\%)}&\scriptsize{\#SV's}&\scriptsize{Time (s)}\\ \hline
{RBP}      &9.44	$\pm$ 0.27&	12624	$\pm$ 0	&188.9&41.45	$\pm$ 0.33&1984 $\pm$ 0&158.2\\
{Forgetron}&9.73	$\pm$ 0.24&	12624	$\pm$ 0	&241.4&41.65	$\pm$ 0.05&1984 $\pm$ 0&227.3\\
{BOGD}     &16.91	$\pm$ 0.05&	12624	$\pm$ 0	&225.2&41.56	$\pm$ 0.02&1984 $\pm$ 0&171.6\\
{BPAS}     &5.62	$\pm$ 0.06&	12624	$\pm$ 0	&239.7&35.09	$\pm$ 0.48&1984 $\pm$ 0&167.6\\

{SPA}      &\textbf{4.34	$\pm$ 0.03}&	12631	$\pm$ 196.5&	\textbf{94.7}&\textbf{34.57	$\pm$ 0.36}	&1978.8	$\pm$ 170.9	&\textbf{70.9}\\
\hline
\end{tabular}
\end{small}
\end{center}\vspace{-0.1in}
\end{table*}
\begin{figure*}[htp]
\begin{center}
\begin{tabular}{ccc}
\includegraphics[width=1.7in]{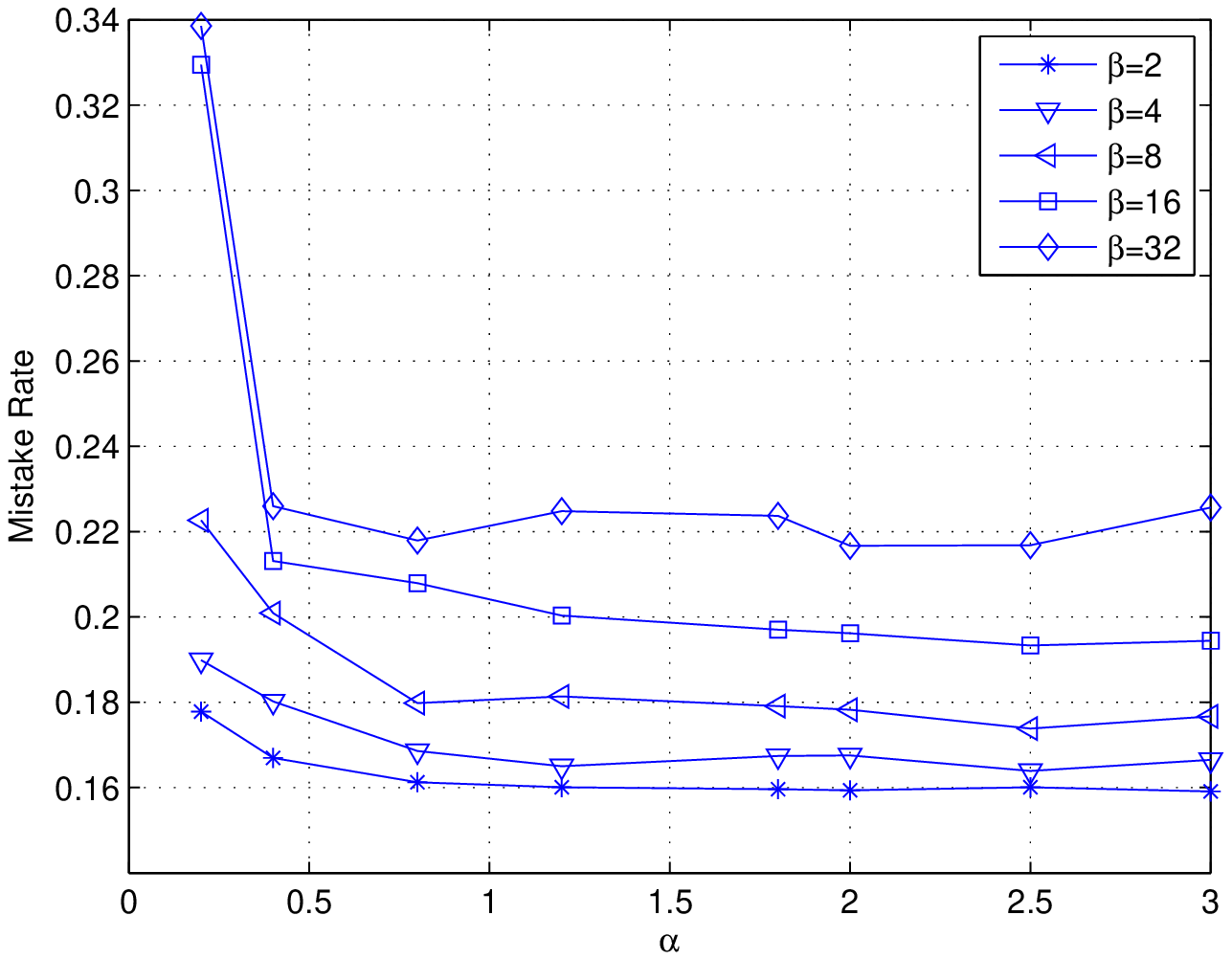}&
\includegraphics[width=1.7in]{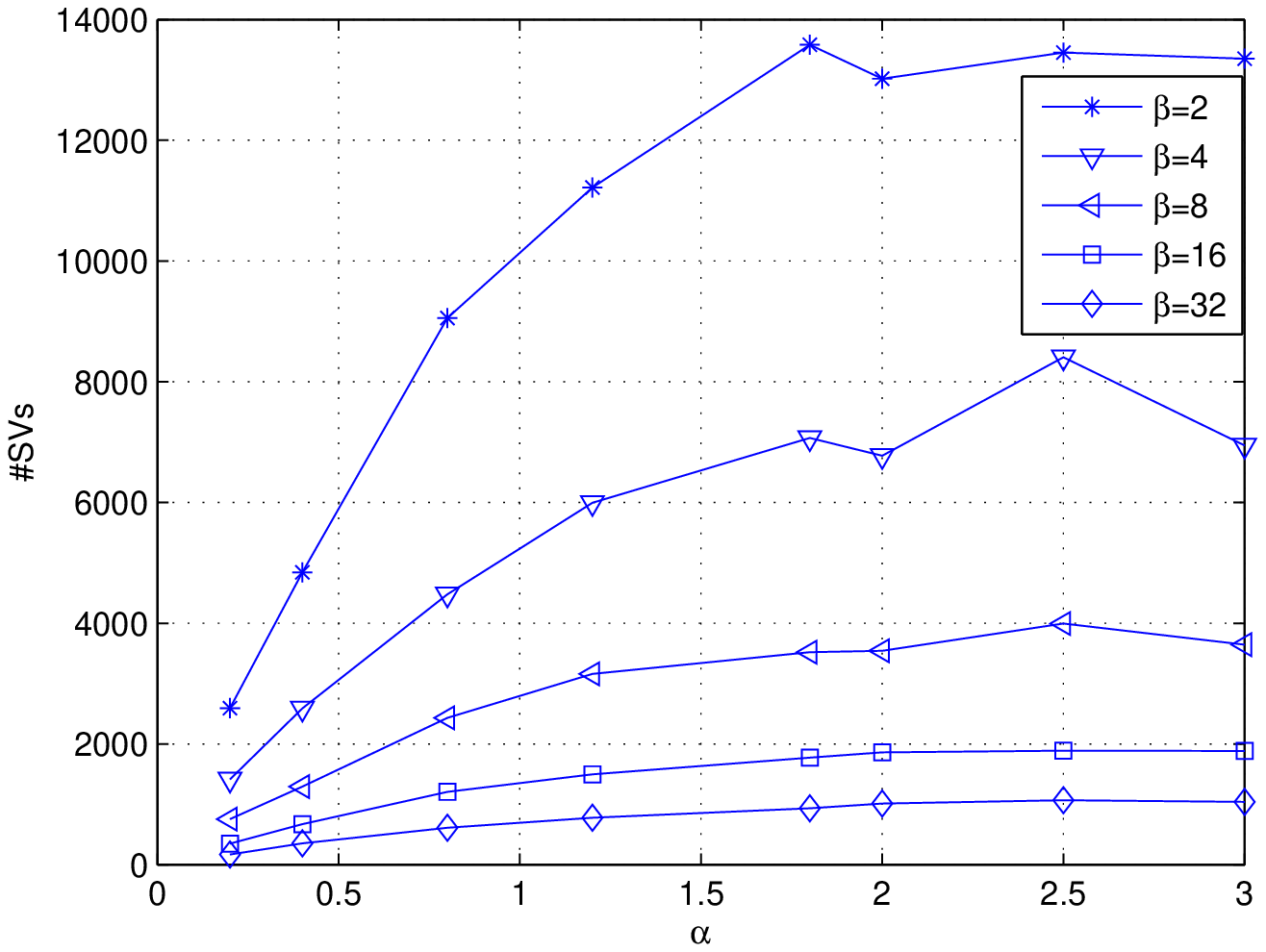}&
\includegraphics[width=1.7in]{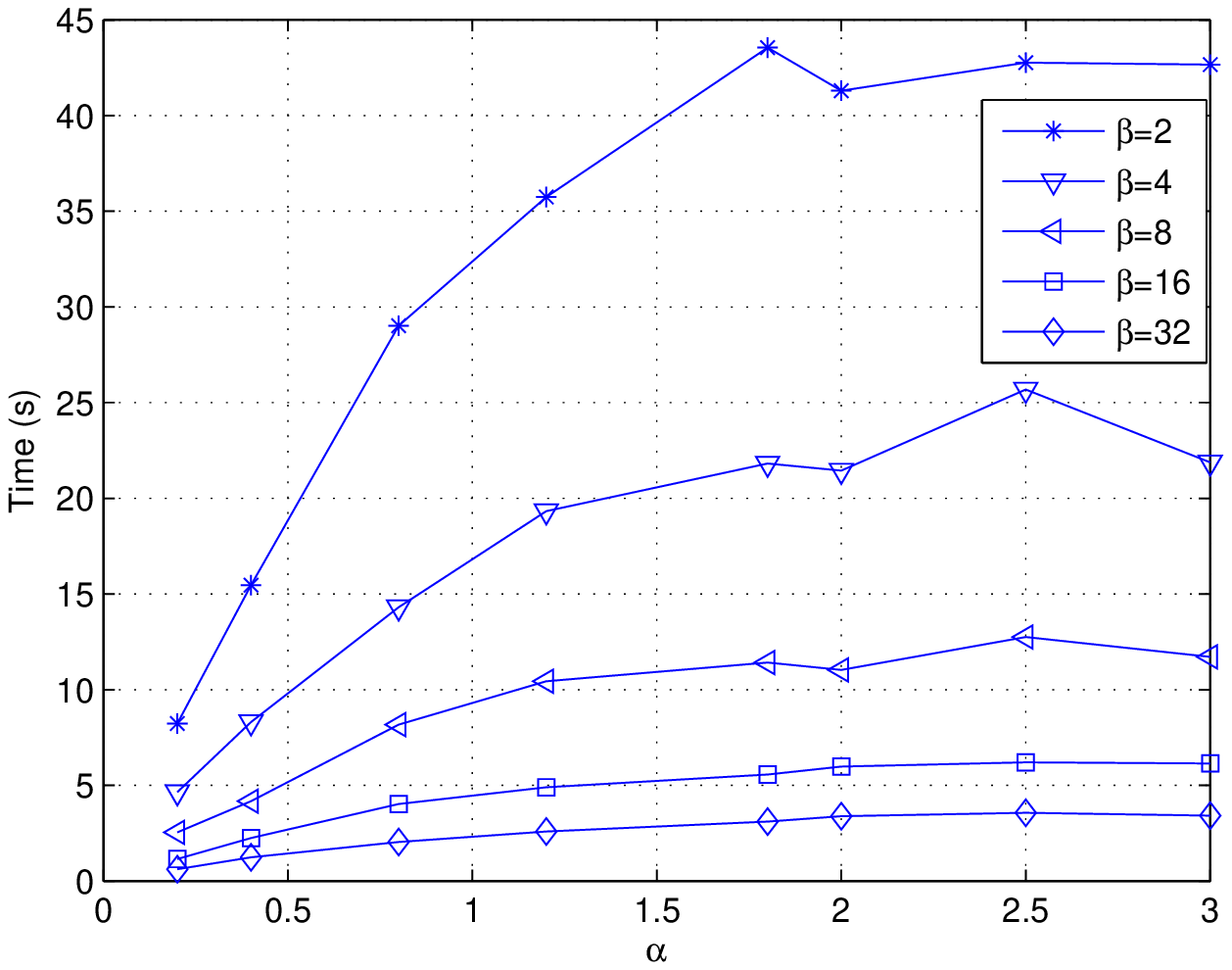}\\
(a) Mistake Rate&
{(b) The number of SV's}&
{(c) Time cost (seconds)}\\
\includegraphics[width=1.7in]{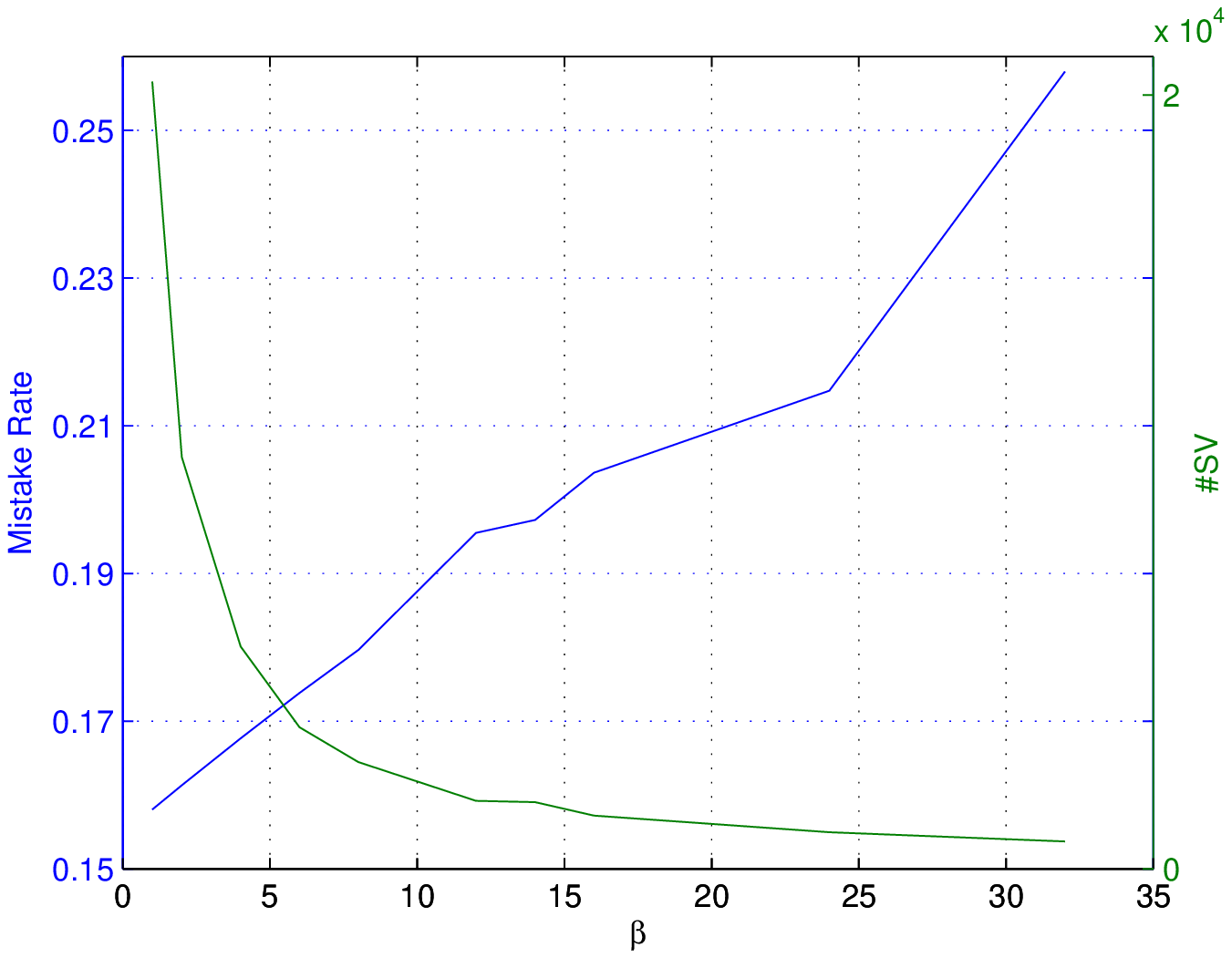}&
\includegraphics[width=1.7in]{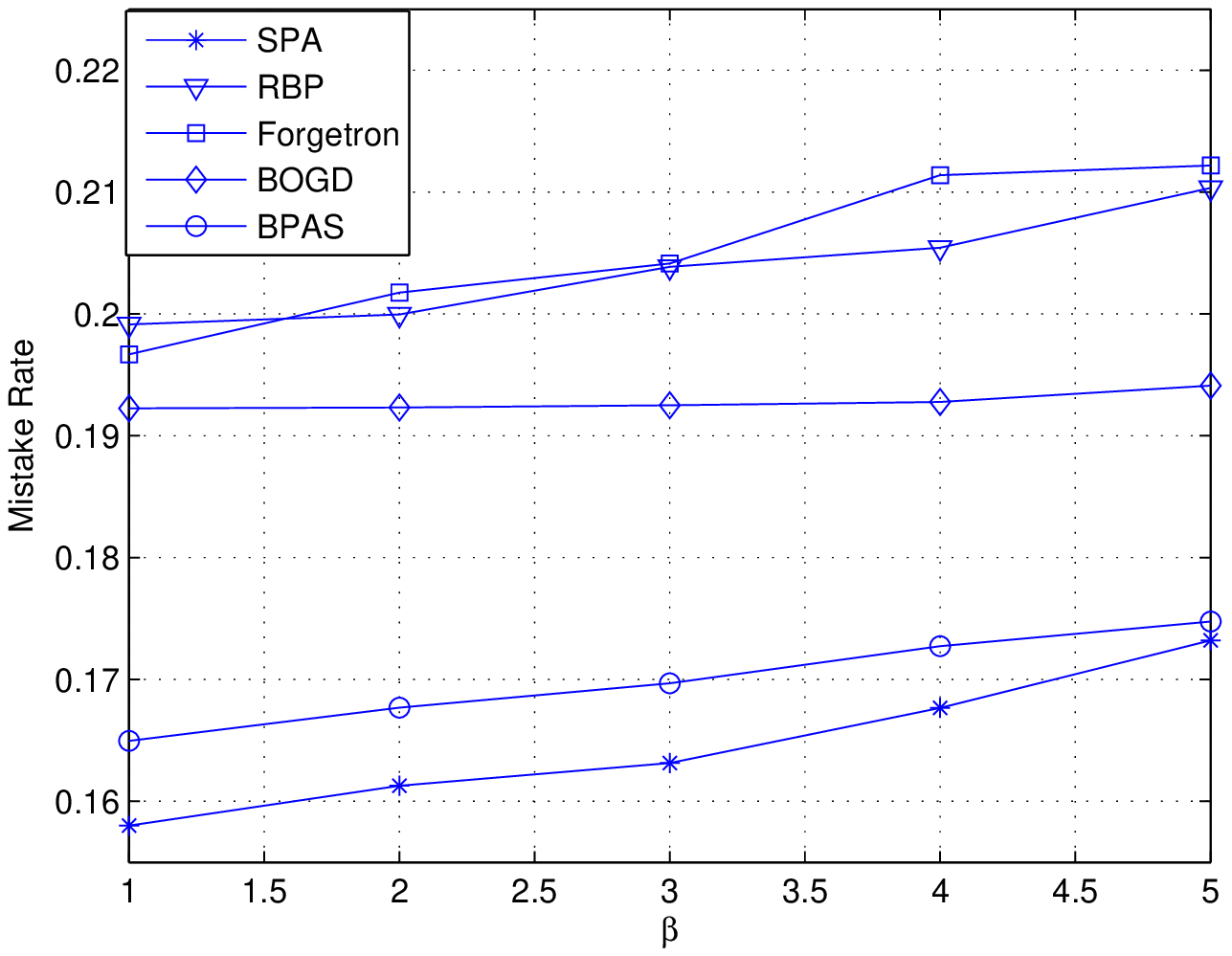}&
\includegraphics[width=1.7in]{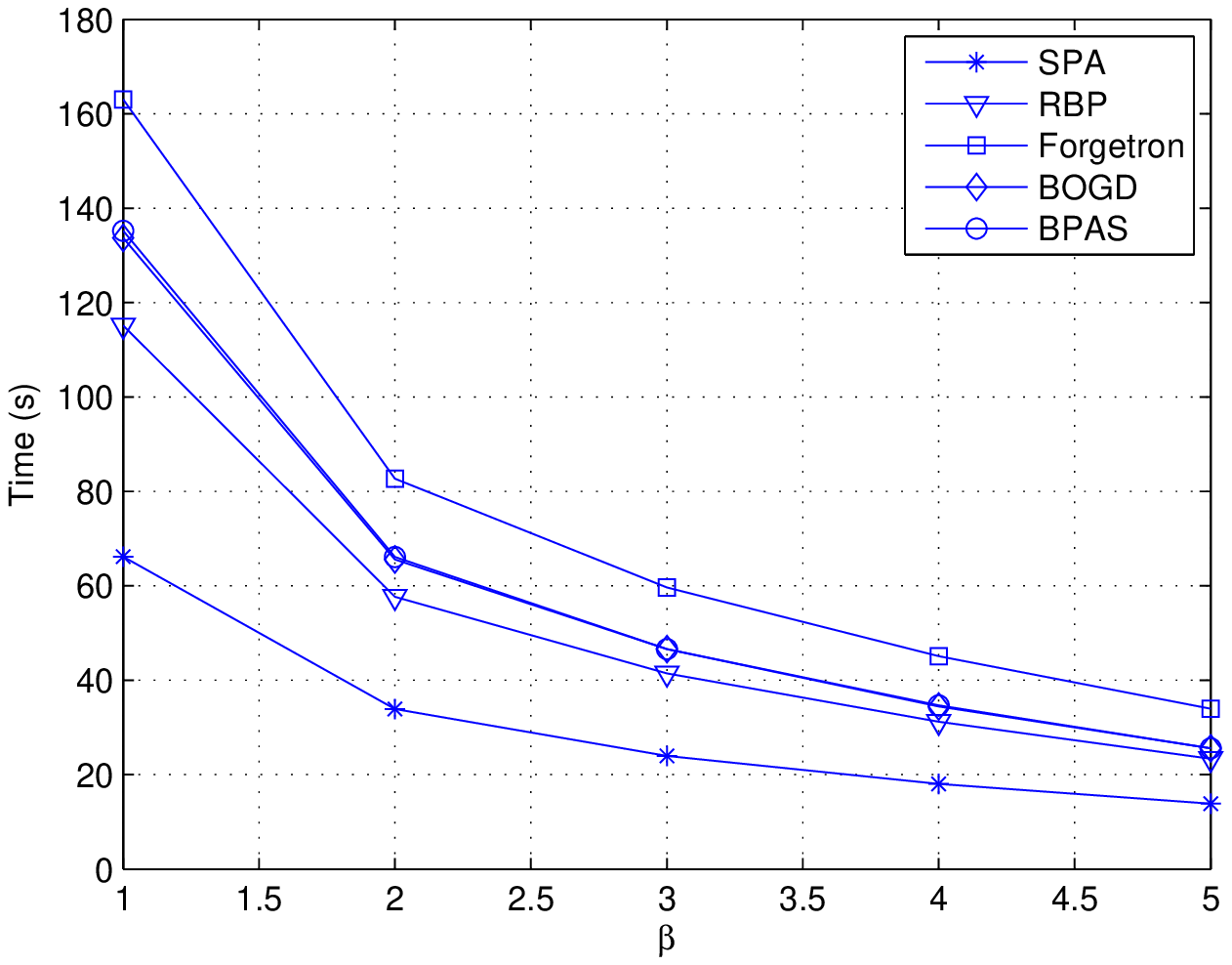}\\
(d) $\beta$, \#SV's and Mistake Rate&
(e) $\beta$ and Mistake Rate&
(f) $\beta$ and Time cost\\
\end{tabular}
\caption{\small{The impact of $\alpha$ and $\beta$ for \#SV's, time cost and mistake rate by the proposed SPA algorithm on dataset ``a9a". We fix $\alpha=1$ for Figure (d), (e) and (f) and examine the impact of $\beta$. For better comparison, we also show the performance of the other budget algorithms in Figure (e) and (f) when using same SV size.}}
\label{icml-historical}
\end{center}
\vspace{-0.2 in}
\end{figure*}
From the results, it is obvious that our SPA algorithm achieves the best accuracy among all the budget algorithms in most of the cases. In addition, the time cost of the proposed SPA algorithm is always the lowest. This is due to the advantage in our algorithm design: SPA can find the optimal kernel and concentrate the effort in this kernel. Thus, poor kernels will receive few SV's and the prediction time is greatly reduced. When adopting the same number of SVs as other budget algorithms, instead of paying equal attention to all components, SPA focuses on the best kernel and thus achieves the highest accuracy.
\subsection{Parameter Sensitivity of $\alpha$ and $\beta$}
The proposed SPA algorithm has two critical parameters $\alpha$ and $\beta$ which could considerably affect the accuracy, support vector size and time cost. Our second experiment is to examine how different parameters of $\alpha$ and $\beta$ affects the learning performance so as to give insights for how to choose them in practice. Figure~\ref{icml-historical} evaluates the performance of the SPA algorithm on the ``a9a" dataset with varied $\alpha$ and varied $\beta$. Several observations can be drawn from the experimental results.

First of all, when $\beta$ is fixed, increasing $\alpha$ generally results in (i) larger support vector size, (ii) higher time cost, but (iii) better classification accuracy, especially when $\alpha$ is small. However, when $\alpha$ is large enough (e.g., $\alpha>1.5$), increasing $\alpha$ has very minor impact to the performance. This is because the number of instances whose hinge loss above $\alpha$ is relatively small. We also note that the accuracy decreases slightly when $\alpha$ is too large. This might be because some (potentially noisy) instances with large loss are given a high chance of being assigned as SV's, which may harm the classifier due to noise. Thus, on this dataset (``a9a"), it is easy to find a good $\alpha$
in the range of [1,2].

Second, when $\alpha$ is fixed, increasing $\beta$ will result in (i) smaller support vector size, (ii) smaller time cost, but (iii) worse classification accuracy. On one hand, $\beta$ cannot be too small as it will lead to too many support vectors and thus suffer very high time cost. On the other hand, $\beta$ cannot be too large as it will considerably decrease the classification accuracy. We shall choose $\beta$ that yields a sufficiently accurate classifier while minimizing the support vector size and training time cost. For example, for this particular dataset, choosing $\beta$ in the range of [3,6] achieves a good trade-off between accuracy and efficiency/sparsity.
\section{Conclusions}
This paper proposed a new framework of Budget Online Budget Multiple Kernel Learning with a novel Sparse Passive Aggressive learning algorithm. In contrast to the existing Online Multiple Kernel Classification (OMKC) algorithms that usually suffer from extremely high time cost due to their unbounded numbers of support vectors in the online learning process, our proposed algorithm adopts a simple but effective SV sampling strategy, making it applicable to large scale applications. We theoretically demonstrated that the SPA algorithm enjoyed an optimal regret bound in expectation. In addition, the experimental results with comparison to both the existing non-budget OMKL algorithms and budget OMKL algorithms showed that the proposed method achieved a significant speedup and yielded more accurate classifiers than existing methods, validating the efficacy, efficiency and scalability of the proposed technique.

{\small
\bibliography{reference1}
\bibliographystyle{iclr2016_conference}
}

\section*{Supplementary Material}
\subsection*{Appendix A: Details of the Datasets used in Experiments}
Table~\ref{dataset} summarizes details of the binary classification datasets used in our experiments. All of them are commonly used benchmark datasets and are publicly available from LIBSVM\footnote{http://www.csie.ntu.edu.tw/~cjlin/libsvmtools/datasets/}, UCI \footnote{http://archive.ics.uci.edu/ml/} and KDDCUP competition site.  These datasets are chosen
fairly randomly to cover a variety of different sizes.
\begin{table}[htpb]
\caption{Statistics of binary classification datasets in our experiments ($T$ denotes the number of instances and $d$ denotes the number of dimensions).} \label{dataset}
\begin{center}
\begin{small}
\begin{sc}
\begin{tabular}{l|ccccccccc}
\hline
&\scriptsize{german}&\scriptsize{svmgudie3}&\scriptsize{madelon}&\scriptsize{magic04}&\scriptsize{a9a}&\scriptsize{kdd08}&\scriptsize{ijcnn1}&\scriptsize{codrna}&\scriptsize{SUSY}\\\hline
T&1000&1,243&2,000&19,020&48,842& 102,294&49,990&271,617 &1,000,000\\
\textnormal{d}&24   &21&500 &10 &123   & 117&22    &8&18\\
\hline
\end{tabular}
\end{sc}
{\scriptsize Note that the original SUSY dataset has more than 5-million instances, from which we randomly sampled one fifth to form our dataset.}
\end{small}
\end{center}
\vspace{-0.1in}
\end{table}
\subsection*{Appendix B: Proof for Lemma 1}
\emph{Proof:}
Firstly, the $P_t(f)$ defined in the equality
\bqs&&\hspace{0.1in}
f_{t+1}=\min_{f\in\Hk}P_t(f):=\frac{1}{2}\|f-f_t\|^2_\Hk+
\frac{Z_t}{\rho_t}\eta \ell_t(f) \eqs  is $1$-strongly convex. Further, $f_{t+1}$ is the optimal solution of $\min_f P_t(f)$, we thus have
the following inequality according the definition of strongly convex
\bqs
\begin{aligned}
\frac{1}{2}\|f-f_t\|^2_\Hk +\frac{1}{\rho_t}Z_t\eta\ell_t(f)\ge\frac{1}{2}\|f_{t+1}-f_t\|^2_\Hk+\frac{1}{\rho_t}Z_t\eta\ell_t(f_{t+1})+\frac{1}{2}\|f-f_{t+1}\|^2_\Hk
\end{aligned}
\eqs
where the inequality used $\nabla P_t(f_{t+1})=0$. After
rearranging the above inequality, we get \bqs\begin{aligned}
&\frac{1}{\rho_t}Z_t\eta\ell_t(f_{t+1})-\frac{1}{\rho_t}Z_t\eta\ell_t(f)\le\frac{1}{2}\|f-f_t\|^2_\Hk
-\frac{1}{2}\|f-f_{t+1}\|^2_\Hk-\frac{1}{2}\|f_{t+1}-f_t\|^2_\Hk\end{aligned}
\eqs Secondly, since $\ell_t(f)$ is $1$-Lipshitz with respect to $f$
\bqs &&\hspace{-0.25in}\ell_t(f_t)-\ell_t(f_{t+1})\le
\|f_t-f_{t+1}\|_\Hk. \eqs Combining the above two inequalities, we
get \bqs &&\hspace{-0.25in}\frac{1}{\rho_t}Z_t\eta
[\ell_t(f_t)-\ell_t(f)]\le\frac{1}{2}\|f-f_t\|^2_\Hk-\frac{1}{2}\|f-f_{t+1}\|^2_\Hk-\frac{1}{2}\|f_{t+1}-f_t\|^2_\Hk+\frac{1}{\rho_t}Z_t\eta
\|f_t-f_{t+1}\|_\Hk \eqs Summing the above inequalities over all $t$
leads to
\bq\label{eqn:cumulative-random-regret}
\sum^T_{t=1}\frac{1}{\rho_t}Z_t\eta\left[\ell_t(f_t)-\ell_t(f)\right]\le\frac{1}{2}\|f-f_1\|^2_\Hk+\sum^T_{t=1}\left[-\frac{1}{2}\|f_{t+1}-f_t\|^2_\Hk+\frac{1}{\rho_t}Z_t\eta
\|f_t-f_{t+1}\|_\Hk\right] \nonumber\eq We now take expectation on the left
side. Note, by definition of the algorithm, $\E_t Z_t= \rho_t$,
where we used $\E_t$ to indicate conditional expectation give all
the random variables $Z_1,\ldots,Z_{t-1}$. Assuming $\rho_t>0$, we
have \bq\label{eqn:rhopositive}
\E\big{[}\frac{1}{\rho_t}Z_t\eta\left[\ell_t(f_t)-\ell_t(f)\right]\big{]}=\E\big{[}\frac{1}{\rho_t}\E_t
Z_t\eta\left[\ell_t(f_t)-\ell_t(f)\right]\big{]}=\eta\E\left[\ell_t(f_t)-\ell_t(f)\right]
\eq Note that in some iterations, $\rho_t=0$, in that case, we have
$\ell_t(f_t)=0$, thus:
 \bq\label{eqn:rhozero1} &&\hspace{-0.5in}\eta[\ell_t(f_t)-\ell_t(f)]\le 0
\eq As mentioned before, $\rho_t=0$ indicates $Z_t=0$ and
$Z_t/\rho_t=0$, we get \bq
&&\label{eqn:rhozero2}\hspace{-0.5in}\frac{1}{\rho_t}Z_t\eta[\ell_t(f_t)-\ell_t(f)]=0
\eq Combining~\eqref{eqn:rhopositive},~\eqref{eqn:rhozero1} and
~\eqref{eqn:rhozero2} and summarizing over all $t$ leads to \bqs
&&\hspace{-0.5in}\eta\E\sum^T_{t=1}\left[\ell_t(f_t)\hspace{-0.05in}-\ell_t(f)\right]\le\E\sum^T_{t=1}\frac{1}{\rho_t}Z_t\eta\left[\ell_t(f_t)-\ell_t(f)\right]
\eqs We now take expectation on the right side of
~\eqref{eqn:cumulative-random-regret}
\begin{small} \bq\label{eqn:tao}\begin{aligned}
\E\left[\frac{1}{2}\|f-f_1\|^2_\Hk\right]+\E\left[\sum^T_{t=1}\left[-\frac{1}{2}\|f_{t+1}-f_t\|^2_\Hk+\frac{1}{\rho_t}Z_t\eta
\|f_t-f_{t+1}\|_\Hk\right]\right]\\
\le\frac{1}{2}\|f\|^2_\Hk+\sum^T_{t=1}\E\left[-\frac{1}{2}\tau_t^2+\frac{1}{\rho_t}Z_t\eta
\tau_t\right]\end{aligned}\eq\end{small} Given all the random variables
$Z_1,\ldots,Z_{t-1}$, we now calculate the conditional expectation
of the variable $M_t=-\frac{1}{2}\tau_t^2+\frac{1}{\rho_t}Z_t\eta
\tau_t$: In probability $\rho_t$, $Z_t=1$ and
$\tau_t=\tau_t'=\min(\frac{\eta}{\rho_t}, \ell_t(f_t))$. We have
$M_{t~(Z_t=1)}=-\frac{1}{2}\tau_t'^2+\frac{1}{\rho_t}\eta \tau_t'$.
And in probability $1-\rho_t$, $Z_t=0$ and $\tau_t=0$. We have
$M_{t~(Z_t=0)}=0$. Considering the two cases, the conditional
expectation is: \bqs\begin{aligned}
\E_t[M_t]&=\rho_tM_{t~(Z_t=1)}+(1-\rho_t)M_{t~(Z_t=0)}=\rho_t\left[-\frac{1}{2}\tau_t'^2+\frac{1}{\rho_t}\eta
\tau_t'\right]<\eta \tau_t'\end{aligned}\eqs In the case when $\alpha \le\ell_t$
and $\rho_t=\frac{\alpha}{\beta}$,
$\tau_t'=\min(\frac{\eta\beta}{\alpha},\ell_t(f_t))\le\frac{\eta\beta}{\alpha}$,
thus $\eta \tau_t'\le \frac{\eta^2\beta}{\alpha}$.

And in the case $\alpha>\ell_t$ and $\rho_t=\frac{\ell_t}{\beta}$,
$\tau_t'=\min(\frac{\eta\beta}{\ell_t(f_t)},\ell_t(f_t))\le\sqrt{\eta\beta}$.
Thus, $\eta \tau_t'\le \frac{\eta^2\beta}{\sqrt{\beta\eta}}$.

Considering both of the cases leads to \bqs \E_t[M_t]<\frac{\eta^2
\beta}{\min(\alpha,\sqrt{\beta\eta})}\eqs

\noindent Summing the above inequality over all $t$ and combining with
\eqref{eqn:tao}, we get\bqs
\eta\E\sum^T_{t=1}\left[\ell_t(f_t)-\ell_t(f)\right]<
\frac{1}{2}\|f\|^2_\Hk + \frac{\eta^2
\beta}{\min(\alpha,\sqrt{\beta\eta})} T \eqs
\noindent Setting
$f=f_*$, and multiplying the above inequality with $1/\eta$ will
conclude the lemma.

\subsection*{Appendix C: Proof for Theorem 1}
\emph{Proof:}
In the following proof, we first generalize the loss bound of the Hedge algorithm \citep{freund1995desicion} to a different situation where 1) stochastic update and stochastic combination are adopted and 2) the hinge loss function we adopt is bounded $\ell_t^i(f)\in[0,L]$ and $L\geq1$. Using the convexity, we have
$$\gamma^{\ell_t^i}\leq 1-\frac{1-\gamma}{\mu}\ell_t^i$$
where $\mu>1$ satisfies the equality when $\ell_t^i=L$ and this constant only depends on $L$ and $\gamma$. We then get
$$\sum_{i=1}^m w^i_{t+1}=\sum_{i=1}^m w^i_t \gamma^{\ell_t^i}\leq \sum_{i=1}^m w^i_t\left(1-\frac{1-\gamma}{\mu}\ell_t^i\right)=(\sum_{i=1}^m w^i_t)\left(1-\frac{1-\gamma}{\mu}\sum_{i=1}^m\theta_t^i\ell_t^i\right)$$
Applying the above formula repeatedly for $t=1,...T$ and using $1+x\leq e^x$ for all real number $x$ yield,
$$\sum_{i=1}^m w_{T+1}^i\leq \exp\left( -\frac{1-\gamma}{\mu}\sum_{t=1}^T\sum_{t=1}^m\theta_t^i\ell_t^i      \right)$$
we may write the above as following,
\bq\sum_{t=1}^T\sum_{t=1}^m\theta_t^i \ell_t^i \leq \frac{-\mu \ln (  \sum_{i=1}^m w_{T+1}^i )  }{1-\gamma}\label{eq:firstinequality}\eq
Obviously, all the weights are positive, thus
$$\sum_{i=1}^m w_{T+1}^i\geq w_{T+1}^i=w_1^i\gamma^{\sum_{t=1}^T \ell_t^i}$$
Plugging this into (\ref{eq:firstinequality}) and replace $w_1^i$ with $1/m$ yields
\bq\sum_{t=1}^T\sum_{t=1}^m\theta_t^i \ell_t^i \leq \frac{-\mu \ln ( w_1^i\gamma^{\sum_{t=1}^T\ell_t^i} )  }{1-\gamma}=\frac{\mu\ln m-\mu\ln \gamma\sum_{t=1}^T\ell_t^i }{1-\gamma}\label{eq:secondinequality}\eq
Using the convexity of loss function, we have
$$\sum_{t=1}^T\ell_t(f_t)=\sum_{t=1}^T\ell_t(\sum_{i=1}^m\theta_t^if_t^i)\leq \sum_{t=1}^T \sum_{i=1}^m\theta_t^i \ell_t^i$$
Thus
\bq\sum_{t=1}^T\ell_t(f_t)\leq\frac{\mu\ln m-\mu\ln \gamma\sum_{t=1}^T\ell_t^i }{1-\gamma}\label{eq:third}\eq
Now we need to bound the accumulate loss of a single classifier, $\sum_{t=1}^T\ell_t^i$. In Lemma 1 we have proven the regret bound of one single kernel classifier learnt by SPA assuming $\E[Z_t]=\rho_t$. Here, we slightly modify this assumption for our proposed SPA multiple kernel classifier, i.e. $\E[Z_t]=\rho_t*p_t^i$, where $\delta<p_t^i<1$. Consequently, the conclusion of Lemma 1 becomes,
\bqs
\delta\E[\sum^T_{t=1}(\ell_t(f_t)-\ell_t(f_*))]<\frac{1}{2\eta}\|f_*\|^2_\Hk+ \frac{\eta \beta}{\min(\alpha,\sqrt{\beta\eta})}T
\eqs
Combining with (\ref{eq:third}) concludes this proof.
\subsection*{Appendix D: Proof for Theorem 2}
\emph{Proof:}
Since $\E_t[Z_t]=\rho_t$, where $\E_t$ is the conditional
expectation, we have \bqs
&&\hspace{-0.25in}\E[\sum^T_{t=1}Z_t]=\E[\sum^T_{t=1}\E_t
Z_t]=\E[\sum^T_{t=1}\rho_t]=\E[\sum^T_{t=1}\min(\frac{\alpha}{\beta}, \frac{\ell_t(f_t)}{\beta})]\le \min(\frac{\alpha}{\beta}T, \frac{1}{\beta}\E\sum^T_{t=1}\ell_t(f_t))\\
&&\hspace{-0.25in}\le \min\left\{\frac{\alpha}{\beta}T,
\frac{1}{\beta}[\sum^T_{t=1}\ell_t(f_*)+
\frac{1}{2\eta}\|f_*\|^2_\Hk+ \frac{\eta
\beta}{\min(\alpha,\sqrt{\beta\eta})}T ]\right\} \eqs which
concludes the first part of the theorem. The second part of the
theorem is trivial to be derived.

\end{document}